\documentclass[letterpaper]{article}
\usepackage{aaai}
\usepackage{times}
\usepackage{helvet}
\usepackage{courier}
\usepackage{xspace}
\usepackage{amssymb}
\usepackage{epic}
\usepackage{graphicx}
\usepackage{subfigure}

{\makeatletter
 \gdef\xxxmark{%
   \expandafter\ifx\csname @mpargs\endcsname\relax 
     \expandafter\ifx\csname @captype\endcsname\relax 
       \marginpar{xxx}
     \else
       xxx 
     \fi
   \else
     xxx 
   \fi}
 \gdef\xxx{\@ifnextchar[\xxx@lab\xxx@nolab}
 \long\gdef\xxx@lab[#1]#2{{\bf [\xxxmark #2 ---{\sc #1}]}}
 \long\gdef\xxx@nolab#1{{\bf [\xxxmark #1]}}
  \long\gdef\xxx@lab[#1]#2{}\long\gdef\xxx@nolab#1{}%
}

\begin{document}
%
\title{Symmetry Breaking Constraints: Recent Results\thanks{Toby 
Walsh is supported by the
Australian Government's Department of Broadband, Communications and
the Digital Economy, the ARC and the Asian Office of Aerospace
Research and Development through grant AOARD-104123
and 124056.
This paper was invited as a "What's Hot" paper to the AAAI'12 Sub-Area 
Spotlights track. }}
\author{Toby Walsh\\ NICTA and UNSW\\ Sydney, Australia\\ toby.walsh@nicta.com.au}

\maketitle
\begin{abstract}
Symmetry is an important problem in many
combinatorial problems. One way of dealing with symmetry
is to add constraints that eliminate
symmetric solutions. We survey recent
results in this area, focusing 
especially on two common and useful 
cases: symmetry breaking 
constraints for row and column symmetry, and 
symmetry breaking constraints for eliminating 
value symmetry. 
\end{abstract}

\section{Introduction}

\newtheorem{mydefinition}{Definition}
\newtheorem{mytheorem}{Theorem}
\newtheorem{mylemma}{Lemma}
\newtheorem{mytheorem1}{Theorem}
\newcommand{\myproof}{\noindent {\bf Proof:\ \ }}
\newcommand{\myqed}{\mbox{$\blacksquare$}}
\newcommand{\myend}{\mbox{$\clubsuit$}}

\newcommand{\mymod}{\mbox{\rm mod}}
\newcommand{\mymin}{\mbox{\rm min}}
\newcommand{\mymax}{\mbox{\rm max}}
\newcommand{\range}{\mbox{\sc Range}}
\newcommand{\roots}{\mbox{\sc Roots}}
\newcommand{\myiff}{\mbox{\rm iff}}
\newcommand{\alldifferent}{\mbox{\sc AllDifferent}}
\newcommand{\permutation}{\mbox{\sc Permutation}}
\newcommand{\pecedence}{\mbox{\sc Precedence}}
\newcommand{\disjoint}{\mbox{\sc Disjoint}}
\newcommand{\cardpath}{\mbox{\sc CardPath}}
\newcommand{\CARDPATH}{\mbox{\sc CardPath}}
\newcommand{\common}{\mbox{\sc Common}}
\newcommand{\uses}{\mbox{\sc Uses}}
\newcommand{\lex}{\mbox{\sc Lex}}
\newcommand{\DLex}{\mbox{\sc DoubleLex}}
\newcommand{\snakelex}{\mbox{\sc SnakeLex}}
\newcommand{\usedby}{\mbox{\sc UsedBy}}
\newcommand{\nvalue}{\mbox{\sc NValue}}
\newcommand{\slide}{\mbox{\sc CardPath}}
\newcommand{\sliden}{\mbox{\sc AllPath}}
\newcommand{\SLIDE}{\mbox{\sc CardPath}}
\newcommand{\circularslide}{\mbox{\sc CardPath}_{\rm O}}
\newcommand{\among}{\mbox{\sc Among}}
\newcommand{\mysum}{\mbox{\sc MySum}}
\newcommand{\amongseq}{\mbox{\sc AmongSeq}}
\newcommand{\atmost}{\mbox{\sc AtMost}}
\newcommand{\atleast}{\mbox{\sc AtLeast}}
\newcommand{\element}{\mbox{\sc Element}}
\newcommand{\gcc}{\mbox{\sc Gcc}}
\newcommand{\gsc}{\mbox{\sc Gsc}}
\newcommand{\contiguity}{\mbox{\sc Contiguity}}
\newcommand{\PRECEDENCE}{\mbox{\sc Precedence}}
\newcommand{\assignnvalues}{\mbox{\sc Assign\&NValues}}
\newcommand{\linksettobooleans}{\mbox{\sc LinkSet2Booleans}}
\newcommand{\domain}{\mbox{\sc Domain}}
\newcommand{\symalldiff}{\mbox{\sc SymAllDiff}}
\newcommand{\alldiff}{\mbox{\sc AllDiff}}
\newcommand{\doublelex}{\mbox{\sc DoubleLex}}

\newcommand{\dyn}{\mbox{\sc Dyn}}
\newcommand{\preced}{\mbox{\sc Prec}}
\newcommand{\prep}{\mbox{\sc Prep}}
\newcommand{\precedsh}{\mbox{\sc $\preced_{sh}$}}
\newcommand{\precedprep}{\mbox{\sc $\preced+\prep$}}
\newcommand{\precedprepsh}{\mbox{\sc $\preced+\prep_{sh}$}}

\newcommand{\slidingsum}{\mbox{\sc SlidingSum}}
\newcommand{\MaxIndex}{\mbox{\sc MaxIndex}}
\newcommand{\REGULAR}{\mbox{\sc Regular}}
\newcommand{\regular}{\mbox{\sc Regular}}
\newcommand{\precedence}{\mbox{\sc Precedence}}
\newcommand{\STRETCH}{\mbox{\sc Stretch}}
\newcommand{\SLIDEOR}{\mbox{\sc SlideOr}}
\newcommand{\NAE}{\mbox{\sc NotAllEqual}}
\newcommand{\mytheta}{\mbox{$\theta_1$}}
\newcommand{\mysigma}{\mbox{$\sigma_2$}}
\newcommand{\mysigmatwo}{\mbox{$\sigma_1$}}

\newcommand{\todo}[1]{{\tt (... #1 ...)}}
\newcommand{\myOmit}[1]{}

\newcommand{\dpsb}{DPSB}

Symmetry occurs in many constraint satisfaction and
optimisation problems \cite{symchap06}.
For example, suppose we have a proper coloring of a graph, 
and we permute the colors then we will obtain another
symmetric coloring. 
Symmetries are problematic as they 
increase the size of the search space. 
We will waste time visiting symmetric solutions.
Worse still, we will waste even more time visiting
the (many failing) parts of the search tree 
which are symmetric to already visited states.
A common and effective method to deal with symmetry is to
add constraints which eliminate some, but not all symmetric
solutions
(e.g. \cite{puget:Sym,clgrkr96,ssat2001,asmijcai2003,pijcai2005,llconstraints06,wcp06}).
In this paper, we survey recent results
on symmetry breaking constraints. 
We hope that this survey is
of wider interest as many of the results are likely to
translate to other domains
like planning, model checking and heuristic
search. 

Symmetry breaking constraints
have a number of good and bad properties.
On the positive side,
a few simple constraints can often eliminate
most if not all symmetry in a problem quickly and
easily. In addition, propagation 
between the problem and the symmetry breaking constraints
can reduce search considerably.
On the negative side, we pick out particular
solutions in each symmetry class, and
this may conflict with the direction
of the branching heuristic.
An alternative to posting static
symmetry breaking constraints is a more
dynamic approach that modifies the search method
to ignore symmetric states
\cite{backofen:Sym,sbds,fahle1,sellmann2,dynamiclex}.
The advantage of a dynamic method
is that it can reduce the conflict
with the branching
heuristic. However, 
we may also get less propagation.
The results reported here on static symmetry breaking
constraints typically have implications for 
dynamic methods as most dynamic methods 
are equivalent to
adding static symmetry breaking 
constraints ``on the fly''. 

There are several aspects of symmetry breaking
that this survey does not cover. 
We do not discuss methods to identify symmetry
in a problem. See, for instance, \cite{pugetcp05,mbwdcpaior08}. 
We suppose that the symmetries are known
in advance as they often are, and our challenge
is merely to deal with them. 
The survey also does not cover other
active area of research like the
intersection of symmetry reasoning and nogood
learning. See, for instance, \cite{bnosictai10,csbmijcai11}.
For background material on constraint satisfaction
in general, see a text like \cite{handbookcp}.

\section{An Example}

We consider a combinatorial problem studied in several
earlier works on symmetry breaking. 

\newenvironment{myexample}{{\bf Running example:} \it}{\rm}

\begin{myexample}
The Equidistant Frequency Permutation Array (EFPA) problem
\cite{hmmncp09} 
is a challenging problem in coding theory. The goal
is to find a set of $v$ code words, each of length $q\lambda$
such that each word contains $\lambda$ copies of the symbols
1 to $q$, and each pair of code words is Hamming distance
$d$ apart. For example, for $v=4$, $\lambda=2$, $q=3$, $d=4$,
one solution is:
\alph{equation}
\renewcommand{\theequation}{\alph{equation}}
\begin{equation} \label{epfa}
\begin{array}{cccccc}
0 & 2 & 1 & 2 & 0 & 1 \\
0 & 2 & 2 & 1 & 1 & 0 \\
0 & 1 & 0 & 2 & 1 & 2 \\
0 & 0 & 1 & 1 & 2 & 2  
\end{array}
\end{equation}
This problem has applications in 
communication theory,
and is 
related to other combinatorial
problems like finding 
orthogonal Latin squares.
We can model this problem by a $v$ by $q\lambda$
array of decision variables with domains $1$ to $q$. 
\end{myexample}

A {\em symmetry} of such a problem is a bijection $\sigma$
that maps solutions onto solutions
\cite{cjjpsconstraints06}.
We consider two special classes of symmetries:
{\em variable symmetries} which act just on variables,
and {\em value symmetries} which act just on values.
Both types of symmetry can be found in our model
of the EFPA problem.

\begin{myexample}
This model of the EFPA problem has variable
symmetry since we can
permute the variables in any two rows,
or in any two columns and still have
a solution. 
For example, we can swap the first column with the third
without changing the frequency of symbols
in each row or the Hamming distance between 
any two rows:
\renewcommand{\theequation}{\alph{equation}}
\begin{equation} \label{epfa2}
\begin{array}{cccccc}
1 & 2 & 0 & 2 & 0 & 1 \\
2 & 2 & 0 & 1 & 1 & 0 \\
0 & 1 & 0 & 2 & 1 & 2 \\
1 & 0 & 0 & 1 & 2 & 2  
\end{array}
\end{equation}
The model also has value symmetry since we can
interchange values throughout any solution. 
For example, the value symmetry $\theta$ swaps the value 0 with 2
throughout solution {\rm (\ref{epfa})}, again
without changing the frequency of symbols
in each row or the Hamming distance between 
any two rows:
\renewcommand{\theequation}{\alph{equation}}
\begin{equation} \label{epfa3}
\begin{array}{cccccc}
2 & 0 & 1 & 0 & 2 & 1 \\
2 & 0 & 0 & 1 & 1 & 2 \\
2 & 1 & 2 & 0 & 1 & 0 \\
2 & 2 & 1 & 1 & 0 & 0  
\end{array}
\end{equation}
\end{myexample}

\section{Symmetry Breaking Constraints}

A simple but effective method to deal with symmetry
is to add constraints which eliminate some
(but not all) symmetric solutions in each
symmetry class.

\begin{myexample}
To eliminate the value symmetry in the 
EFPA problem, we can add 
a {\sc Precedence} constraint 
\cite{llcp2004} to ensure that the values
in the bottom row occur in order. That is,
the value 0 first occurs on the bottom row before 1, 
and 1 first occurs before 2.
The bottom row of solution {\rm (\ref{epfa})}
is $0 0 1 1 2 2$ which satisfies
this symmetry breaking constraint. 
The value 0 first occurs in the first position,
the value 1 later on in the third position, and
the value 2 even later on in the fifth position.
Suppose we swap 0 with 2, then
the bottom row of solution {\rm (\ref{epfa})} becomes $2 2 1 1 0 0$,
the bottom row of {\rm (\ref{epfa3})}.
This violates the {\sc Precedence}
symmetry breaking constraint. The 
value 1 first occurs in the third position,
which is before the first occurrence of the
value 0 in the fifth position. Hence, by
posting this symmetry breaking constraint,
we only find solution {\rm (\ref{epfa})},
and the symmetric solution {\rm (\ref{epfa3})} 
is eliminated.
\end{myexample}

Two important properties of symmetry
breaking constraints are soundness and completeness.
A set of symmetry
breaking constraints is {\em sound} iff it
leaves at least one solution in each symmetry
class, and {\em complete}
iff it leaves at most one solution in each symmetry
class. 
The {\sc Precedence} constraint is, for example, sound and complete. 
Where do symmetry breaking constraints like this come 
from in general?
The Lex-Leader method offers
a generic method for deriving a sound and complete set of
symmetry breaking constraints 
for variable symmetries \cite{clgrkr96}
and value symmetries \cite{wcp07}.
The method picks out the lexicographically smallest solution in each
symmetry class. For every symmetry $\sigma$,
it posts a lexicographical ordering constraint:
$$ \langle X_1, \ldots, X_n \rangle \leq_{\rm lex}
\sigma (\langle X_1, \ldots, X_n \rangle)
$$
Where $X_1$ to $X_n$ is some fixed ordering on the
variables in the problem. 

\begin{myexample}
Consider again the value symmetry, $\theta$ that
swaps 0 and 2. We can describe this by the mapping 
$\theta(X)=2-X$. The Lex-Leader method
eliminates this symmetry with the 
ordering constraint:
$$ \langle X_1, \ldots, X_{24} \rangle \leq_{\rm lex}
\langle 2-X_1, \ldots, 2-X_{24} \rangle
$$
Where $X_1$ to $X_{24}$ is some ordering of the
24 decision variables modelling our 4 by 6 instance
of the EFPA problem.
This simplifies to:
$$ \langle X_1, \ldots, X_{24} \rangle \leq_{\rm lex}
\langle 1, \ldots, 1 \rangle
$$
Suppose we order the variables in the matrix row-wise 
from top left to bottom right.
Then this constraint would
accept solution {\rm (\ref{epfa})},
but eliminate solutions {\rm (\ref{epfa2})} and
{\rm (\ref{epfa3})}.
\end{myexample}

Many static symmetry breaking constraints
can be derived from such Lex-Leader constraints.
For example, 
{\sc Precedence} constraints
to break the symmetry due to interchangeable
values can be
derived from them \cite{wecai2006}.
Efficient algorithms have
been developed to propagate
lexicographical ordering
constraints \cite{fhkmwcp2002,fhkmwaij06,knwercim09,bnqwcpaior11}.

\section{Symmetries of Symmetry Breaking Constraints}

One way of constructing new symmetry breaking
constraints is to use symmetry itself \cite{kwecai10}. 
Any symmetry acting on a set of symmetry breaking 
constraints will itself break the symmetry in a problem. 
This requires us to 
define the action of a symmetry on a set of symmetry 
breaking constraints. We defined symmetry as acting on assignments, 
mapping solutions to solutions. We can
lift this definition to constraints. For example,
the action of a variable symmetry on a constraint changes the 
scope of the constraint (that is, 
the variables on which the constraint acts). 

\begin{mytheorem}[\cite{kwecai10}]
Given a set of symmetries $\Sigma$ of C, if S is a 
sound (complete) set of symmetry breaking constraints for 
$\Sigma$ then $\sigma(S)$ for any $\sigma \in \Sigma$
is also a sound (complete) set of symmetry breaking constraints for
$\Sigma$. 
\end{mytheorem}

Different symmetries pick out different solutions in 
each symmetry class. In fact, if we have a particular
solution in mind, we can pick it out using
a suitable symmetry of a set of symmetry
breaking constraints. Let $sol(C)$ be the set of solutions
of a set of constraints $C$. 

\begin{mytheorem}[\cite{kwecai10}]
Given a set of symmetries $\Sigma$ of a set of constraints C, 
a sound set
S of symmetry breaking constraints, and any solution A of C, then
there is a symmetry $\sigma \in \Sigma$
such that $A \in sol(C \cup \sigma(S))$. 
\end{mytheorem}

Applying symmetry to a set of symmetry breaking
constraints does not change the symmetries which are eliminated.
We say that a set of constraints $S$ 
{\em breaks} a symmetry 
$\sigma$ of a problem $C$ iff there exists a solution $A$ of $C \cup S$ 
such that $\sigma(A)$ is not a solution of $C \cup S$,
and {\em completely breaks} a symmetry 
$\sigma$ iff for each solution $A$ of $C \cup S$,
$\sigma(A)$ is not a solution of $C \cup S$. 
Given a symmetry group $\Sigma$, we say that a set of 
constraints (completely) breaks $\Sigma$ iff it breaks every
non-identity symmetry in $\Sigma$. 

\begin{mytheorem}[\cite{kwecai10}]
Given a problem $C$ and a symmetry group $\Sigma$, 
if $S$ (completely) breaks $\Sigma$
then $\sigma(S)$ (completely) breaks $\Sigma$
for any $\sigma \in \Sigma$.
\end{mytheorem}

We have used these ideas as the theoretical 
basis for a hybrid symmetry breaking method that 
tries to have the best of both worlds, posting static
symmetry breaking constraints dynamically during
search according to the direction of the branching
heuristic \cite{kwecai10}. 

\section{Intractability of Breaking Symmetry}

The symmetry groups seen in practice can be very large.
As a result, symmetry breaking can be computationally
challenging. 
For example, if we have a $q$ interchangeable 
values (as in our EFPA model),
then we have $q!$ symmetries. 
To eliminate each of these symmetries requires
a separate Lex-Leader constraint. 
As a consequence, the Lex-Leader method is
not tractable in general.
An alternative is to break symmetry partially 
by posting just a subset of the Lex-Leader
constraints (e.g. \cite{jpcp11}). 

Crawford {\it et al.} 
(\citeyear{clgrkr96}) proved
that breaking symmetry completely by adding constraints
to eliminate symmetric solutions
is computationally intractable in general.
More specifically, they prove that, given a matrix
model with row and column
symmetries, deciding if an assignment is the
smallest in its symmetry class is NP-hard where
we append rows together
and compare them lexicographically.
There is, however, nothing
special about appending rows together or
comparing assignments lexicographically. We
could break symmetry with {\em any} total ordering
over assignments.

Recently we have shown that, under
modest assumptions, breaking symmetry
remains intractable if we use a different 
ordering of variables, or even 
a different ordering over solutions
than the lexicographical ordering. 
More precisely, we show that the problem
is NP-hard for any {\em simple} ordering
where, given an assignment, 
we can compute the position in the
ordering in polynomial time and,
in the reverse, given a position 
in the ordering, we can compute
the associated assignment in polynomial
time. 

\begin{mytheorem}[\cite{wcp11}]
Given any simple ordering, there 
exists a symmetry group such that deciding if an assignment is 
smallest in its symmetry class according to this ordering is NP-hard.
\end{mytheorem}

Since breaking symmetry appears intractable
in general, a major research direction is 
to identify special cases which occur in practice
where the symmetry group is more tractable to break. 
We consider two such cases: row and column symmetry,
and value symmetry.

\section{Row and Column Symmetry}

A matrix of decision variables has {\em row symmetry} iff 
given a solution, any permutation of the rows is
also a solution.
Similarly, it has {\em column symmetry} iff 
given a solution, any permutation of the columns is
also a solution.
For example, our model of the EFPA problem has 
both row and column symmetry. 
Row and column symmetry occurs in many
models with matrices of decision variables
\cite{lex2001,matrix2001,ffhkmpwcp2002}. 
We can break row and and column
symmetry using the Lex-Leader method. However,
this is not practical in general.
A $n$ by $m$ matrix has $n!m!$ row and column 
symmetries, and each symmetry would require
a separate lexicographical ordering constraint. 

To break all row symmetry, we can
post a linear number of constraints
that lexicographically order
the rows. Similarly,
to break all column symmetry we can 
post a linear number of constraints
that lexicographically order
the columns. 
When we have both row and column symmetry,
we can post 
a $\DLex$ constraint that 
lexicographical orders
both the rows and columns
\cite{ffhkmpwcp2002}.
In fact, $\DLex$ constraints
can be derived from a subset of the constraints introduced
by the Lex-Leader
method \cite{lex2001}.
As a result, $\DLex$ does not break
all row and column symmetry.
Nevertheless, it is often highly effective in practice. 

\begin{myexample}
Consider again solution 
\mbox{\rm (\ref{epfa})}.
If we order the rows of \mbox{\rm (\ref{epfa})} lexicographically, we 
get a solution in which rows and columns
are ordered lexicographically:
\renewcommand{\theequation}{\alph{equation}}
\begin{equation} \label{epfa22}
\begin{array}{cccccc}
0 & 2 & 1 & 2 & 0 & 1 \\
0 & 2 & 2 & 1 & 1 & 0 \\
0 & 1 & 0 & 2 & 1 & 2 \\
0 & 0 & 1 & 1 & 2 & 2  
\end{array}
\
\begin{array}{c}
\mbox{\rm order} \\
\Rightarrow \\
\mbox{\rm rows}
\end{array}
\ \
\begin{array}{cccccc}
0 & 0 & 1 & 1 & 2 & 2 \\
0 & 1 & 0 & 2 & 1 & 2 \\
0 & 2 & 1 & 2 & 0 & 1 \\
0 & 2 & 2 & 1 & 1 & 0 
\end{array}
\end{equation}
Similarly 
if we order the columns of \mbox{\rm (\ref{epfa})} lexicographically, we 
get a different solution in which both rows and columns
are again ordered lexicographically:
\renewcommand{\theequation}{\alph{equation}}
\begin{equation} \label{epfa33}
\begin{array}{cccccc}
0 & 2 & 1 & 2 & 0 & 1 \\
0 & 2 & 2 & 1 & 1 & 0 \\
0 & 1 & 0 & 2 & 1 & 2 \\
0 & 0 & 1 & 1 & 2 & 2  
\end{array}
\ 
\begin{array}{c}
\mbox{\rm order} \\
\Rightarrow \\
\mbox{\rm cols}
\end{array}
\ \ 
\begin{array}{cccccc}
0 & 0 & 1 & 1 & 2 & 2 \\
0 & 1 & 0 & 2 & 1 & 2 \\
0 & 1 & 2 & 0 & 2 & 1 \\
0 & 2 & 2 & 1 & 1 & 0  
\end{array}
\end{equation}

All three solutions are thus in the same row and column symmetry 
class. Note that both 
\renewcommand{\theequation}{\alph{equation}}
\mbox{\rm (\ref{epfa22})} and 
\mbox{\rm (\ref{epfa33})}
satisfy the $\DLex$ constraint.
Therefore $\DLex$ can leave multiple
solutions in each symmetry class and is
not a complete symmetry breaking method. 
\end{myexample}

In fact, $\DLex$ is a very incomplete method for breaking
symmetry. 
It can leave $n!$ symmetric solutions in an $2n$ by $2n$ matrix
model. 
In addition, it only partially provides tractability.
Whilst it is polynomial to check
a $\DLex$ constraint given a complete
assignment, it is not tractable to propagate it
completely given a partial assignment. That is,
pruning all symmetric values is NP-hard. 

\begin{mytheorem}[\cite{knwcp10}]
Propagating the \mbox{$\DLex$} constraint is 
NP-hard. 
\end{mytheorem}

We have identified three tractable cases 
where we can break all row and column
symmetry in polynomial time \cite{knwcp10}:
(1) a matrix with a bounded number of rows (or columns);
(2) 
a 0/1 matrix model where
every row sum is 1;
(3) an all-different matrix where 
all entries are different.
The first two are perhaps the most interesting
and useful cases. We have, for example, used
the first case as the basis of a complete method
for breaking row and column symmetry. We used this 
to measure how many 
symmetries are left by \DLex\ in
practice. See Table 1 for some results that
demonstrate \DLex\ actually leaves very few symmetric
solutions in practice despite the worst case result that
it can leave a factorial number of such solutions. More recently, 
Yip and Van Hentenryck \shortcite{yhijcai11}
have turned this theoretical result into a complete 
and efficient method for breaking all row and column symmetry in
matrix models with 
a small number of rows (or columns).
Another interesting research direction
is to identify other constraints 
that can be effectively added to \DLex\ to increase
the amount of row and column symmetry broken \cite{fjmcp2003}.

\begin{table}[htb]
\begin{center}
{\scriptsize
\begin{tabular}{|r||c|c|c|c|}
\hline
$(q, \lambda, d, v)$& \# symmetry
& \# symmetric
&{$\DLex$}
\\ 
 & 
 classes&
 solutions &
 \# solutions / time \\
\hline 
$ (3, 3,2,3) $  &  $\mathbf{6}$ &  $ 1.81{\cdot} 10^5 $    &     $ \mathbf{6} $  / $ {0.00} $   \\
$ (4, 3,3,3) $  &  $8$ &   $ > 3.88{\cdot} 10^7 $    &     $ {16} $  / $ {0.01} $  \\
$ (4, 4,2,3) $  &  $\mathbf{12}$ &   $ > 5.87{\cdot} 10^7 $    &     $ \mathbf{12} $  / $ {0.00} $  \\
$ (3, 4,6,4) $  &  $1427$ &   $ > 5.57{\cdot} 10^7 $    &     $ 11215 $  / $ 5.88 $  \\
$ (4, 3,5,4) $  &  $8600$ &   $ > 2.03{\cdot} 10^7 $    &     $ 61258 $  / $ 69.90 $  \\
$ (4, 4,5,4) $  &  $9696$ &   $ > 5.45{\cdot} 10^6 $    &     $ 72251 $  / $ 173.72 $  \\
$ (5, 3,3,4) $  &  $5$ &   $ > 4.72{\cdot} 10^6 $    &     $ {20} $  / $ 0.36 $  \\
$ (3, 3,4,5) $  &  $18$ &   $ > 2.47{\cdot} 10^7 $    &     $ 71 $  / $ 0.17 $  \\
$ (3, 4,6,5) $  &  $4978$ &   $ > 2.08{\cdot} 10^7 $    &     $ 77535 $  / $ 167.50 $   \\
$ (4, 3,4,5) $  &  $441$ &   $ > 6.55{\cdot} 10^6 $    &     $ 2694 $  / $ 19.37 $  \\
$ (4, 4,2,5) $  &  $\mathbf{12}$ &   $ > 6.94{\cdot} 10^6 $    &     $ \mathbf{12} $  / $ 0.02 $   \\
$ (4, 4,4,5) $  &  $717$ &   $ > 6.27{\cdot} 10^6 $    &     $ 4604 $  / $ 38.15 $  \\
$ (4, 6,4,5) $  &  $819$ &   $ > 4.08{\cdot} 10^6 $    &     $ 5048 $  / $ 69.83 $  \\
$ (5, 3,4,5) $  &  $3067$ &   $ > 2.39{\cdot} 10^6 $    &     $ 20831 $  / $ 403.97 $  \\
$ (6, 3,4,5) $  &  $15192$ &   $ > 2.16{\cdot} 10^6 $    &     $ 1.11{\cdot} 10^5 $  / $ 4924.41 $  \\
\hline 
\end{tabular}}
\caption{\label{t:t1} Equidistant Frequency Permutation Array problems. 
Number of solutions left when posting \DLex\ symmetry breaking constraints. 
 }
\end{center}
\end{table}

\section{SnakeLex}

An interesting alternative method for
breaking row and column symmetry uses
a lexicographical ordering which linearizes the matrix 
in a snake-like manner instead of row-wise. 
This method appends the
first row to the reverse of the second
row, and this is appended to the third
row, and then the reverse of the fourth
row, and so on. 
Breaking 
row and column symmetry with this ordering
is intractable, as it is with the 
more conventional lexicographical ordering where 
matrices are ordered row-wise. 

\begin{mytheorem}[\cite{nwcp12}]
Deciding if an assignment is 
smallest in the snake-lex ordering 
up to row and column permutation is NP-hard.
\end{mytheorem}

The $\snakelex$ constraint,
which is derived from the Lex-Leader
method using such a snake-lex ordering, has been
shown to a promising alternative to 
$\DLex$ \cite{snakelex}. 
To break column symmetry, $\snakelex$ ensures that the first column is 
lexicographically smaller than or equal to
both the second and third columns,
the reverse of the second column
is lexicographically smaller than or equal to
the reverse of both the third and fourth columns,
and so on up till the penultimate column
is compared to the final column.
To break row symmetry, $\snakelex$ ensures that each neighbouring pair
of rows, $X_{1,i},\ldots,X_{n,i}$
and $X_{1,i+1},\ldots,X_{n,i+1}$ satisfy the entwined
lexicographical ordering:
\begin{eqnarray*}
\langle X_{1,i}, X_{2,i+1}, X_{3,i}, X_{4,i+1}, \ldots \rangle \leq_{\rm lex} & & \\
~ ~ ~ ~ ~ ~ ~ ~ ~ \langle X_{1,i+1}, X_{2,i}, X_{3,i+1}, X_{4,i}, \ldots \rangle
\end{eqnarray*}
$\DLex$ breaks the subset of the row and column
symmetries that swap neighbouring rows and columns.
$\snakelex$, by comparison, breaks a strict 
superset of these symmetries. Not surprisingly, it
is a little more expensive to propagate but this 
greater cost is often worthwhile in pruning search. 
Like $\DLex$, $\snakelex$ is also an incomplete
symmetry breaking method that can leave an exponential
number of symmetric solutions \cite{knwcp10}. 

\section{Gray Code Ordering}

Another interesting alternative method for
breaking symmetry is 
to eliminate all but the smallest symmetric
solution in some {\em other} ordering
than the lexicographical ordering.
For example, we have seen some promise in
using the Gray code ordering \cite{nwcp12}. 
The Gray code ordering is a total ordering over
assignments used in error correcting codes. The
4-bit Gray code orders assignments as follows:
\begin{eqnarray*}
&0000,
0001,
0011,
0010,
0110,
0111,
0101,
0100,& \\
&\ \ \ \ \ \ \ \ 1100,
1101,
1111,
1110,
1010,
1011,
1001,
1000
&
\end{eqnarray*}
Such an ordering may pick out different
solutions in each symmetry class. This can
reduce the conflict between 
symmetry breaking, the problem constraints
and the branching heuristic. 
The Gray code ordering has some properties that
may make it useful for symmetry breaking. For example,
neighbouring assignments in the ordering only differ
at one position, and flipping one bit reverses
the ordering of the subsequent bits.
Note that there is no renaming of the variables
and values 
that maps the ordering of solutions in the Gray code
ordering onto that of the lexicographical ordering.
Nevertheless, breaking row and column symmetry
with this ordering over solutions remains
intractable. 

\begin{mytheorem}[\cite{nwcp12}]
Deciding if an assignment is 
smallest in the Gray code ordering 
up to row and column permutation is NP-hard.
\end{mytheorem}

Experiments suggest
some promise for 
symmetry breaking methods using
the Gray code ordering 
\cite{nwcp12}. 
See Table 2 for some example results 
on a benchmark domain, 
the maximum density still life problem.
This is prob032 in CSPLib \cite{csplib}.
Given a $n$ by $n$ sub-matrix of the
infinite plane, we want to find
the maximum density pattern which does
not change from generation to generation.
This problem has the 8 symmetries of the square as
we can rotate or reflect any still life to
obtain a new one.
We broke all symmetries in the problem 
with either the lexicographical
or Gray code orderings, finding
the smallest (lex, gray) or largest
(anti-lex, anti-gray) solution in each
symmetry class. 

\begin{table}[htbp]
{\scriptsize
\begin{center}
\begin{tabular}{|c||r|r|r|r|r|} \hline
{Symmetry breaking} & 
$n=4$ & 5 & 6 & 7 & 8 \\ \hline \hline
 none & 
 176 & 1,166 & 12,205 & {231,408} & 5,867,694 \\  
 gray & 
 91 & 446 & 5,702 & 123,238 & 2,507,747 \\
 anti-lex & 
 84 & 424 & 5,473 & 120,112 & 2,416,266 \\
 lex & 
 33 & 406 & 2,853 & 87,781 & 1,982,698 \\
 anti-gray & 
 {26} & 269 & 2,288 & 38,476 & 1,073,659 \\
\hline
\end{tabular}
\end{center}
}
\caption{Backtracks required to find and prove 
optimal the maximum density still life of size $n$ by $n$. 
}
\end{table}

\section{Value Symmetry}

A second important class of symmetries which are 
more tractable to break than symmetries in general
are value symmetries.  
When a problem has many value symmetries,
the Lex-Leader method again introduces too 
many symmetry breaking constraints to
be practical. Puget has proposed an effective
alternative that maps value symmetries
into variables symmetries  \cite{pcp05}.
His method channels into some auxiliary variables
which, as in \cite{sswaaai2000}, permit
the (symmetry breaking) constraints to be easily
specified. $Z_j$ records the
first index using each value $j$ with constraints
of the form:
\begin{eqnarray*}
X_i = j & \Rightarrow & Z_j \leq i \\
Z_j = i & \Rightarrow & X_i = j
\end{eqnarray*}
Puget then breaks
the variable symmetry on the $Z_j$ by posting binary
ordering constraints. For example,
if the values 1 to $q$ are interchangeable,
we can strictly order the $Z_j$:
\begin{eqnarray*}
& Z_1 < Z_2 < Z_3 < \ldots < Z_q & 
\end{eqnarray*}
This ensures that the first occurrence of 1 is
before that of 2, that of 2 is before
3, etc. This is the condition enforced by the {\sc Precedence}
constraint mentioned previously.
Puget proved that we can break
{\em any} number of value symmetries (and not just
those due to interchangeable values) with a linear
number of ordering constraints on the $Z_j$.
This means that we can break any
number of value symmetries in polynomial
time. 
Unfortunately, 
this decomposition into ordering 
constraints on $Z_j$ hinders propagation,
and we may not prune all
symmetric values. 
This is not surprising as
pruning all symmetric values
is intractable in general. 

\begin{mytheorem}[\cite{wcp07}]
Given a set of value symmetries and
a partial assignment, pruning all symmetric values
is NP-hard. 
\end{mytheorem}

We have, however, identified a common case where
pruning all symmetric values is tractable. 
If values partition into a 
fixed number of sets and 
values within each set are interchangeable
then we can prune all symmetric
values in polynomial time 
using a global {\sc Regular} constraint
\cite{pesant1} 
or one of its decompositions
\cite{qwcp06,qwcp07}. 
When there is a single set of interchangeable
values, we can prune all symmetric
values using a global {\sc Precedence} constraint 
\cite{llcp2004}
or its decomposition
\cite{wecai2006}. 

\section{Combinations of Symmetries}

Another important 
question is how we deal with with combinations
of symmetries. We have already seen
that we can take a symmetry group that 
is tractable to break completely (row symmetry),
combine it with another symmetry group that 
is tractable to break completely (column symmetry),
and end up with a symmetry group (row and column
symmetry) that is intractable to break and
where we lack good methods to eliminate {\em all}
symmetric solutions in general. 

Another challenging combination is when
we have both variable and value symmetries.
How do we break both types of symmetry 
simultaneously? We can use the Lex-Leader method
but this is not practical when we have many
variable and value symmetries. Unfortunately,
Puget's method for breaking
value symmetry on the $Z_j$ variables is not
compatible in general with breaking
variable symmetry via the Lex-Leader method on
the $X_i$ variables. 
This corrects Theorem 6 and Corollary 7 in \cite{pcp05}
which claim that the two methods are compatible if
we use the same ordering of variables in each method. 

\begin{mytheorem}[\cite{knwcp10}]
There exist problems on
which posting lex-leader constraints
to break variable symmetries and applying Puget's method 
to break value symmetries remove all solutions
in a symmetry class irrespective of the orderings
on variables used by both methods.
\end{mytheorem}

There is, however, an important and 
common case where
the two methods for breaking variable
and value symmetry are compatible. If 
values partition into interchangeable sets
then it is always consistent to post
symmetry breaking constraints using both
methods provided we index their variables 
identically in both cases
\cite{sellmann2,fpsvcp06,llwycp07}. 
This is again a case where tractability 
is limited as pruning
all symmetric values is NP-hard.

\section{Open Problems}

A number of important 
and challenging questions about 
symmetry breaking constraints remain
including:
\begin{description}
\item[Identifying tractable cases:] Are there other common
types of symmetry which occur in practice 
that are polynomial to break? For instance,
suppose we are coloring the edges of a graph, 
and we have a clique of interchangeable vertices.
This induces a particular symmetry on the
decision variables representing the colors of 
the different edges in the graph. How do we efficiently break
this large symmetry group?
\item[Exploring other orderings:] Are there other
orderings besides the lexicographical ordering
with which we can break symmetry effectively? 
We have seen some promise for the Gray code
ordering. There are, however, many other 
orderings we could consider. 
For example, Frisch {\em et al.} have
proposed the multiset ordering \cite{fhkmwijcai2003}.
Even though this is only a partial ordering, it
has also shown some promise. 
In addition, can we 
use a different ordering in each symmetry class?
\item[Reducing branching conflict:] One of the major
issues with symmetry breaking constraints
is that they may conflict with the direction
of the branching heuristic. How can we 
retain the benefits of static symmetry
breaking constraints and of dynamic branching
heuristics, whilst avoiding such conflict? 
How do we get the best of static and dynamic
symmetry breaking methods? How do we choose
static symmetry breaking constraints that align
with a dynamic branching heuristic?
\item[Studying combinations of symmetries:]
How do we eliminate combinations of
symmetries effectively? There are
issues of both soundness (which symmetry
breaking constraints can be combined
together safely?), completeness 
(how we do eliminate all combinations of
symmetry?) and tractability (which symmetry groups
that are tractable to break individually combine
together to give a symmetry group which
is also tractable to break?).
\item[Dominance detection and elimination:]
in constraint optimisation problems, the
notion of dominance plays a very similar
role to symmetry. One (partial) solution dominates
another if it is sure to be at least as good
in quality. How do we identify dominating 
solutions? Is there a generic method like Lex-Leader
for adding constraints that
remove dominated solutions? What are 
interesting and useful tractable cases?
\end{description}



\end{document}